\documentclass[pmlr,twocolumn,10pt]{jmlr} 

\usepackage{booktabs}
\usepackage{makecell}
\usepackage{subcaption}
\jmlrvolume{LEAVE UNSET}
\jmlryear{2023}
\jmlrsubmitted{LEAVE UNSET}
\jmlrpublished{LEAVE UNSET}
\jmlrworkshop{Machine Learning for Health (ML4H) 2023} 

\title[A Comparative Analysis of Machine Learning Models]{A Comparative Analysis of Machine Learning Models\\for Early Detection of Hospital-Acquired Infections}

\author{
  \Name{Ethan Harvey}$^{1,2}$\Email{ethan.harvey@tufts.edu}\\
  \Name{Junzi Dong}$^1$\Email{junzi.dong@gmail.com}\\
  \Name{Erina Ghosh}$^1$\Email{erina.ghosh@philips.com}\\
  \Name{Ali Samadani}$^1$\Email{a.a.samadani@gmail.com}\\
  \addr$^1$Philips Research North America, Cambridge, MA, USA\\
  \addr$^2$Department of Computer Science, Tufts University, Medford, MA, USA\\
}

\begin{document}

\maketitle

\begin{abstract}
As more and more infection-specific machine learning models are developed and planned for clinical deployment, simultaneously running predictions from different models may provide overlapping or even conflicting information.
It is important to understand the concordance and behavior of parallel models in deployment.
In this study, we focus on two models for the early detection of hospital-acquired infections (HAIs): 1)~the Infection Risk Index (IRI) and 2)~the Ventilator-Associated Pneumonia (VAP) prediction model.
The IRI model was built to predict all HAIs, whereas the VAP model identifies patients at risk of developing ventilator-associated pneumonia.
These models could make important improvements in patient outcomes and hospital management of infections through early detection of infections and in turn, enable early interventions.
The two models vary in terms of infection label definition, cohort selection, and prediction schema.
In this work, we present a comparative analysis between the two models to characterize concordances and confusions in predicting HAIs by these models.
The learnings from this study will provide important findings for how to deploy multiple concurrent disease-specific models in the future.

\end{abstract}

\section{Introduction}
Hospital-acquired infections (HAIs) are nosocomial infections that are acquired after hospitalization and start at least 48 hours after hospital admission.
HAIs include central line-associated bloodstream infections, catheter-associated urinary tract infections, surgical site infections, hospital-acquired pneumonia, ventilator-associated pneumonia, and clostridium difficile infections \citep{ducel2002prevention}.
In addition to adverse effects on patient health, HAIs impose a significant financial burden on the healthcare system, adding upward of \$40,000 in cost per hospital stay \citep{zimlichman2013health}.

In this work, we present a comparative analysis between two HAI prediction models: 1)~the Infection Risk Index (IRI), and 2)~the Ventilator-Associated Pneumonia (VAP) model.
IRI was developed as an early warning predictive algorithm capable of alerting medical personnel to individuals at a high risk of HAIs, whereas the VAP model estimates individualized risk of ventilator-associated pneumonia ahead of clinical suspicion of infection \citep{samadani2022vap}.
It is unclear how this infection-specific model, trained on a hospitalized patient subpopulation (i.e., mechanically-ventilated patients) compares to IRI and ultimately, how the two models can be deployed together.
We hypothesize models trained to predict all HAIs (such as IRI) suffer from non-specific predictions and that clinical decision support systems suffer from missingness inherent in patient records.

\section{Materials and Methods}
To align the IRI and VAP models for the comparative analysis, we use the same prediction schema including sample definition, observation window, prediction gap, and prediction frequency for both models.
We use a common test cohort with the aligned prediction schema to compare the performance of the two models and analyze whether false positives and/or false negatives of the VAP prediction model are indicative of other HAIs.

There are major differences between the two models in terms of infection labeling. The IRI model relies on a confirmed infection diagnosis via International Classification of Diseases (ICD) codes while the VAP prediction model relies on clinical actions to label suspected VAP events. In particular, for IRI, the infection cohort consists of adults patients ($\ge$18y) with a confirmed HAI diagnosis via ICD codes and presence of either a culture order or a non-prophylactic antibiotic treatment in health records. Furthermore, patients with culture or non-prophylactic antibiotics within the first 48 hours of ICU admission were excluded. 
For the VAP model, a new antibiotic treatment temporally contiguous to a culture order constitutes a suspected infection event. A suspected infection event is then labeled as presumed VAP if 1)~the culture returns positive, 2)~the event occurs at least 48 hours after intubation, and 3)~there are no indications of community-acquired pneumonia (CAP) or other HAIs in the patient charts \citep{american2005guidelines,center2002ventilator}. For both models, the onset time of the infection is determined by the time of antibiotics’ administration or culture, whichever is earlier.

For the comparative analysis, we use a one-shot sampling approach with a prediction gap of 24 hours, resulting in one sample per patient. We use feature specific observation windows: 12 hours for vitals, 24 hours for ventilator settings, 24 hours for ventilator-associated event (VAE)~\citep{center2002ventilator} related features, and 24 hours for labs. When a feature is missing in an observation window, we search the temporal interval immediately preceding the observation window and use the most recent measurement for the feature if it exists, where vital signs are valid for 2 hours, ventilator settings for 2 hours, VAE-related features are valid for 2 hours, and labs are valid for 26 hours.

We compare the methods used by the IRI and VAP models for masking missingness and clinical concerns from the model, namely, 1)~Gaussian random sampling imputation and 2)~stratified sampling to balance missingness rate between classes (balanced class-specific missingness).
For Gaussian random sampling imputation, we randomly sample from a Gaussian distribution $\mathcal{N}(\mu,\sigma)$ defined on the normal reference range $(l, u)$ with $\mu = (l+u)/2$, $\sigma = 0.15*(u-l)$.
For balancing class-specific missingness, we implement stratified sampling anchored at the temperature measurement in training patients to ensure that temperature is missing uniformly between the two classes.
We select temperature as it is frequently measured and an important vital sign for infection mornitoring.
Ensuring equal class-specific missingness rate for temperature results in more similar rates of missingness in all other features.

Unlike IRI, the VAP prediction model includes model safeguards to control, as feasible, the irrelevant sources of variations due to length of stay (LOS). The VAP prediction model subsamples control patients (non-VAP patients) so that the LOS is matched between the two classes (VAP and non-VAP). This case-control matching is done through stratified sampling of non-VAP patients so that the distribution of LOS for these patients matches that of the VAP patients. The LOS matching helps to avoid a selection bias situation where the model learns higher LOS as a risk of VAP (even if LOS is not used as a feature) as typically non-VAP patients have much shorter LOS as compared with VAP patients.

\textbf{Dataset:} We compare the IRI and VAP models for the early detection of HAIs using electronic health record (EHR) data from the eICU Collaborative Research Database~\citep{pollard2018eicu}. We use the overlap between the cohorts of these models as a common test cohort (Figure~\ref{fig:Venn_Diagram}). Appendix~\ref{apd:Common_Test_Cohort} shows the distribution of labels in the common test cohort.

\begin{figure}[htbp!]
\floatconts
  {fig:Venn_Diagram}
  {\caption{The overlap of ICU stays between the patient cohorts used in the IRI and VAP models.}}
  {\includegraphics[width=0.75\linewidth]{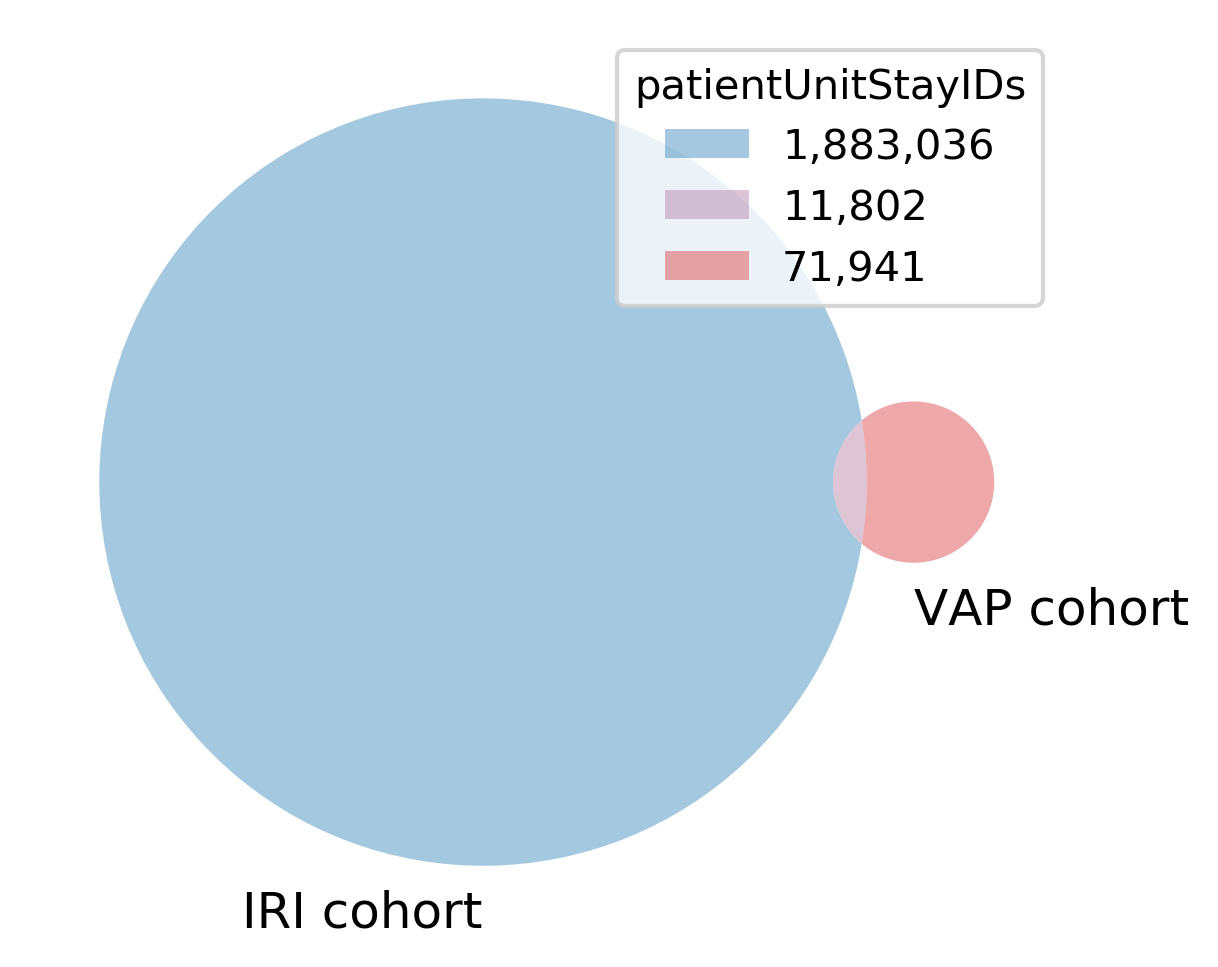}}
\end{figure}

\begin{table*}[t!]
\floatconts
  {tab:Imputation_Methods}
  {\caption{Training, validation, and testing performance of the IRI and VAP models with varying imputation methods.}}
  {\small\begin{tabular}{lccc}
  \toprule
  \bfseries Model & \bfseries \makecell{Training\\AUC ROC} & \bfseries \makecell{Validation\\AUC ROC} & \bfseries \makecell{Testing\\AUC ROC} \\
  \midrule
  IRI using Gaussian random sampling imputation & $0.84 \pm 0.01$ & $0.83 \pm 0.01$ & $0.58 \pm 0.01$ \\
  IRI with balanced class-specific missingness & $0.83 \pm 0.00$ & $0.82 \pm 0.00$ & $0.60 \pm 0.01$ \\
  VAP using Gaussian random sampling imputation & $0.79 \pm 0.01$ & $0.75 \pm 0.01$ & $0.72 \pm 0.01$ \\
  VAP with balanced class-specific missingness & $0.79 \pm 0.03$ & $0.73 \pm 0.01$ & $0.81 \pm 0.01$ \\
  \bottomrule
  \end{tabular}}
\end{table*}

The test patients could be very different in their characteristics from the training patients of the models and can be regarded as out of distribution (OOD) samples.
To mitigate the OOD problem, we add $1/5$ of the common test cohort to the training set and we use the remaining $4/5$ of the common test cohort to compare inter model performance.
We repeat all experiments with five different train, validation, and test sets, ensuring each patient's data belongs to exactly one set.

For both models, we trained an ensemble of extreme gradient boosting (Xgboost) trees~\citep{chen2016xgboost}.
We selected hyperparameters that maximized the area under the receiver operator characteristic (AUC ROC) curve for an internal validation set extracted from the training set.

\section{Results}

Table~\ref{tab:Imputation_Methods} shows the training, validation, and testing performance of the IRI and VAP models with varying imputation methods.
Although the VAP prediction model outperforms IRI on the common test cohort, it is unclear if predicting all HAIs is more difficult than predicting VAP.

We include confusion matrices for positive and negative samples from the VAP model. We found that with the threshold at the Youden's index $(\theta = 0.52)$, 20\% of false positives from the VAP prediction model were patients with other HAIs (see Figure~\ref{fig:Test_Matrices}).

\begin{figure}[htbp!]
\floatconts
  {fig:Test_Matrices}
  {\caption{Confusion matrices for positive and negative samples from the VAP prediction model using balancing class-specific missingness.}}
  {\includegraphics[width=1\linewidth]{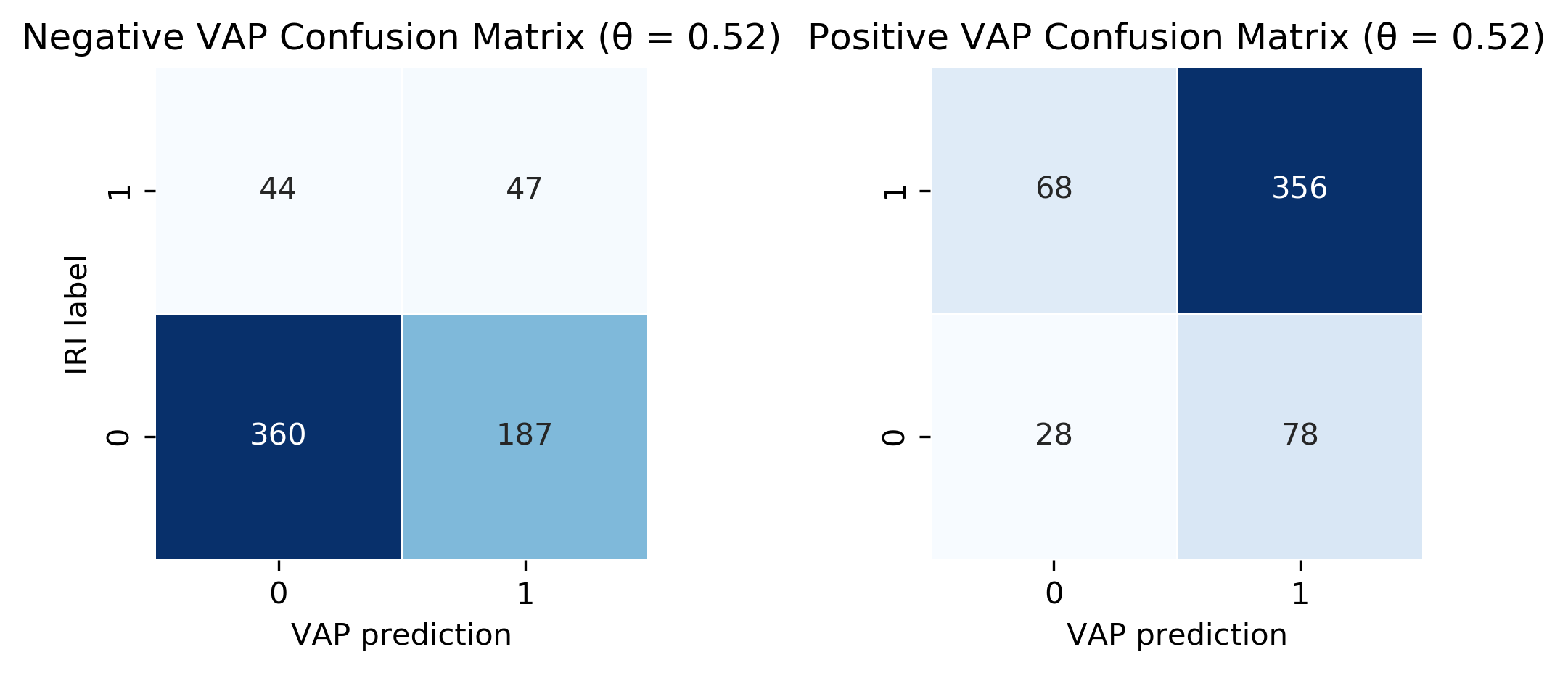}}
\end{figure}

To further evaluate the VAP prediction model's performance, we compared its performance on the IRI and VAP labels (see Figure~\ref{fig:Test_AUCs}).
Although the VAP prediction model is detecting other HAIs and performs better than IRI models for MV patients, further investigation needs to be done to evaluate using the VAP prediction model for early detection of all HAIs in MV patients.

\begin{figure}[htbp!]
\floatconts
  {fig:Test_AUCs}
  {\caption{The performance of the VAP prediction model with balanced class-specific missingness on the IRI and VAP labels. We averaged the receiver operator characteristic (ROC) curves from each test set by taking vertical samples of the ROC curves for fixed false positive rates and averaging the corresponding true positive rates. Confidence intervals of the mean true positive rate were computed using the common assumption of a binomial distribution \citep{fawcett2006introduction}.}}
  {\includegraphics[width=1\linewidth]{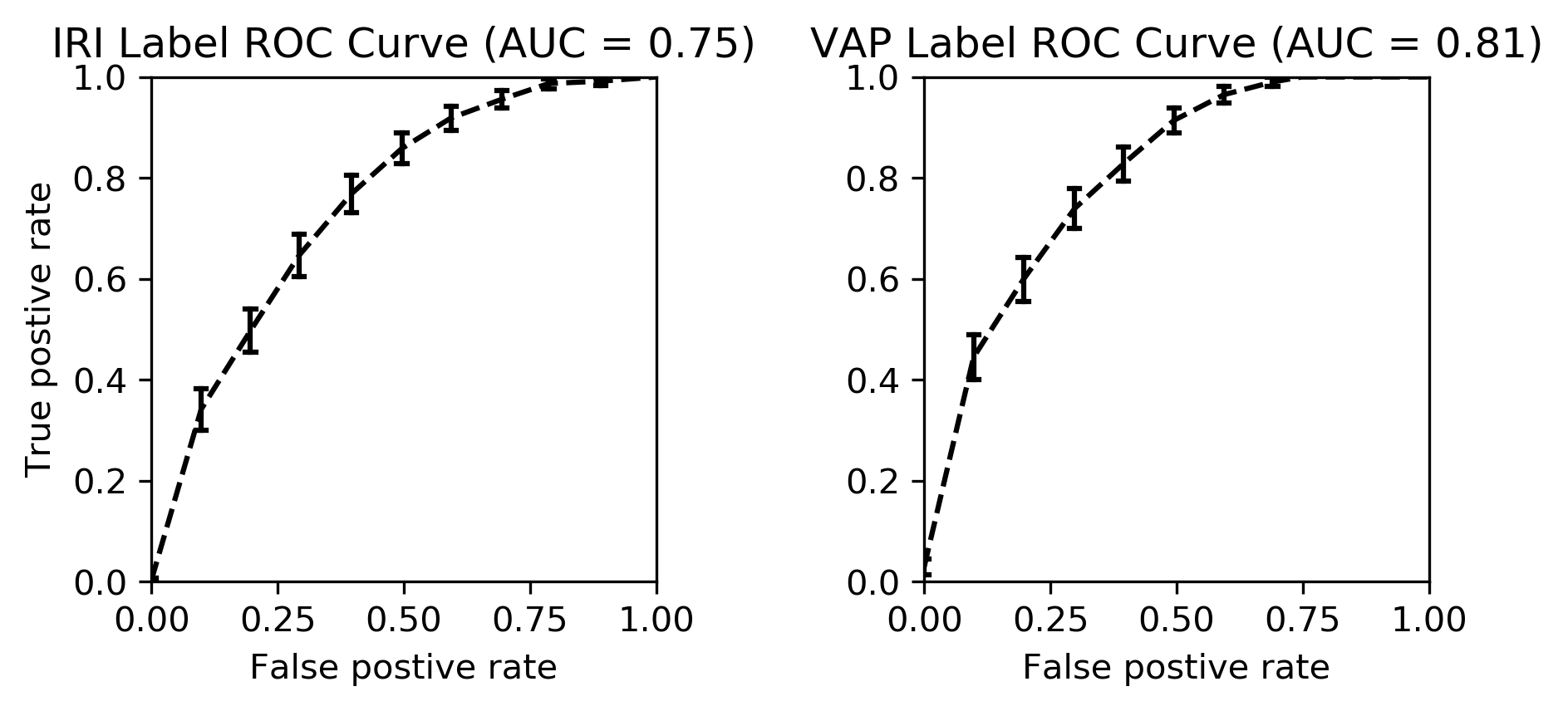}}
\end{figure}

\section{Discussion}
We found that features relying on ventilator settings (e.g., \texttt{SpO2FiO2Ratio}) had high SHAP (SHapley Additive exPlanations \citep{lundberg2020local}) values for IRI models.
The high importance of these features results in significant missingness since only 1\% of the IRI patient cohort is mechanically ventilated.
We found that IRI underperforms on the MV patient subpopulation.
Therefore, we propose that instead of simultaneously running predictions from different models, the VAP prediction model be used if a patient is mechanically ventilated.

We recommend more attention be directed towards infection label definitions.
A common approach to identify adverse events or diagnoses in EHR is to use ICD codes.
This approach does not provide the onset time of the events, as ICD codes are typically assigned at the discharge point only.
Even if limited to a one-shot prediction at a pre-set time (e.g., 48hrs after intubation \citep{giang2021predicting}), the prediction time might overlap with the infection window (temporal window where an HAI is present and has already been clinically observed).
More importantly, ICD codes are not a reliable indicator of adverse events during patient stays, particularly ICD codes used to chart VAP diagnosis \citep{wolfensberger2018should}.
We recommend that models rely on clinical actions to label suspected infection events as was done for the VAP model.

Furthermore, we recommend more attention be directed towards curating the study cohort using approaches such as propensity score matching \citep{rosenbaum1983central} to create similar control and case groups, in order to safeguard the downstream predictive models against selection biases including missingness patterns, LOS, and other variables describing patient phenotypes and hospital subgroups.
For instance, the absence of a lab or culture order is likely due to the lack of clinical concern, which results as missing data in the patient charts. 

The Gaussian imputation is based on a strong assumption that the underlying distribution of the feature is normal and will result in regression to mean for the imputed values, which is susceptible to carrying the missingness pattern in the form of similar values for all the imputed cases.
We found that features relying on the length of MV  (e.g., \texttt{mv\_hrs}) had high SHAP values when Gaussian random sampling imputation was used for the VAP prediction model.
Associating longer episodes with a higher risk of VAP is a  selection bias learned by the model.
Balancing class-specific missingness ensures the model is not associating any inherent class-specific missingness patterns with the risk of infection.

The IRI model does not subsample patients so that the LOS matches between classes (see Figure~\ref{fig:HAI_Event_Day}).
It is critical to ensure that the control class is matched with the case class in terms of LOS, as was done in the VAP model, to avoid a selection bias situation where the LOS is a predictive feature, associating longer episodes with a higher risk of HAIs.

\begin{figure}[htbp!]
\floatconts
  {fig:HAI_Event_Day}
  {\caption{IRI LOS comparison for control and infected patients.}}
  {\includegraphics[width=0.75\linewidth]{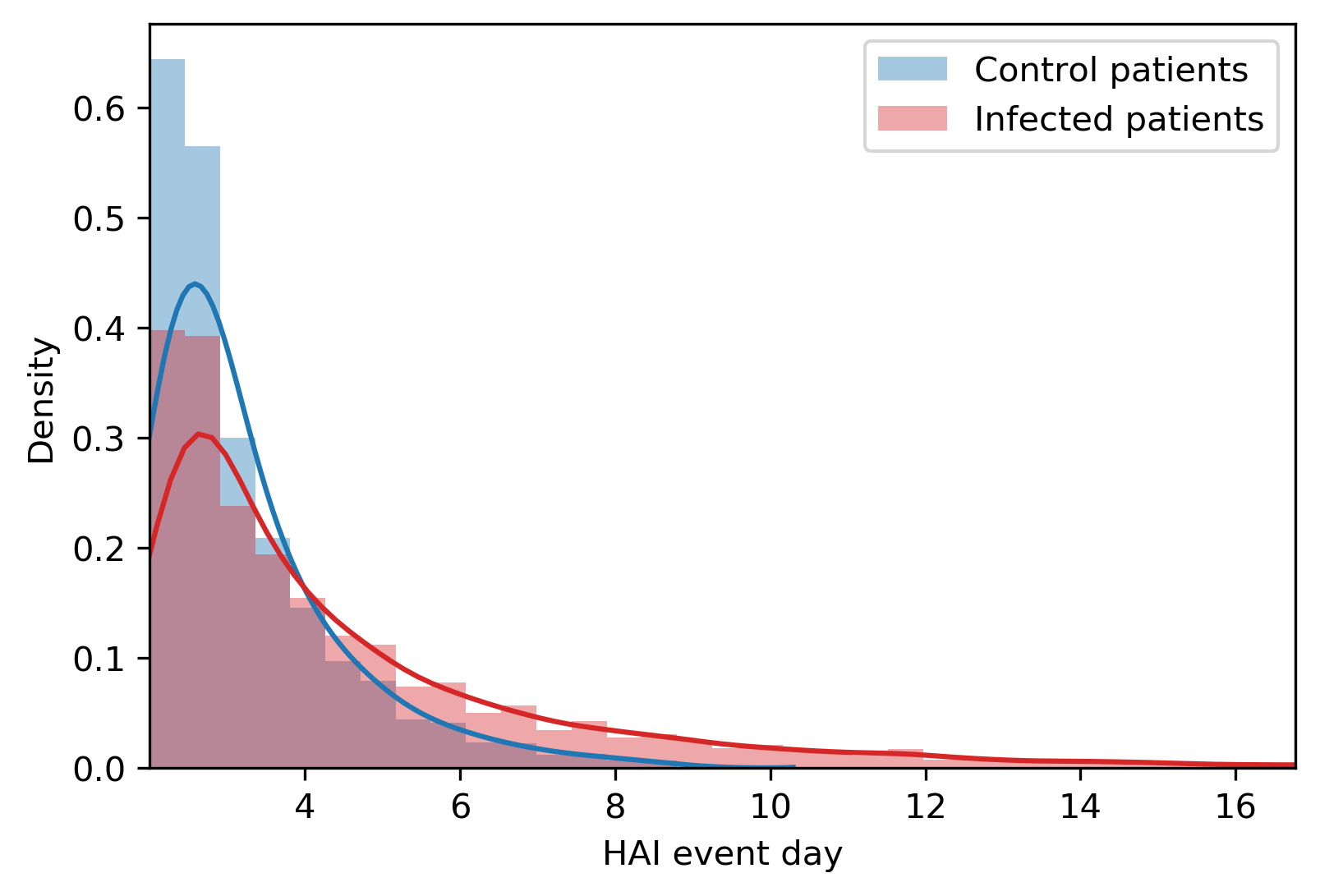}}
\end{figure}

Finally, the IRI model drops all patients who had culture or non-prophylactic antibiotics in the first 48 hours of their stay.
This reduces the number of common patients shared in the two studies, the number of infected patients in the IRI model, and excludes a harder positive class, patients who were suspected to have an infection upon entering the ICU and developed an infection later during their stay.
We propose relaxing the constrain on no cultures or non-prophylactic antibiotics taken in the first 48 hours of ICU stay as was used in IRI model. 

\section{Conclusion}
We presented a comparative analysis between two HAI prediction models in a scenario where the two models generate predictions for the same patients at the same point in time.
We found both models may struggle from selection bias introduced by class-specific missingness in clinical variables and imbalanced case-control groups for features indicating severity (e.g., LOS).
In our experiments, the models with balanced class-specific missingness and matched case-control groups performed better on unseen test patients.
The model trained to predict all HAIs  underperforms on mechanically ventilated patients.
We proposed that instead of simultaneously running predictions from different models, the VAP prediction model be used if a patient is mechanically-ventilated.
We recommend more attention be directed towards infection label definitions and curating the study cohort to safeguard the downstream models against selection biases including length of stay and those resulting from clinical concerns.

\bibliography{harvey23}

\appendix

\section{Common Test Cohort}
\label{apd:Common_Test_Cohort}

\begin{table}[htbp]
\floatconts
  {tab:Label_Distribution}
  {\caption{Distribution of labels in the common test cohort.}}
  {\small\begin{tabular}{lc}
  \toprule
  \bfseries Label & \bfseries \makecell{Percent\\of cohort} \\
  \midrule
  Positive IRI and positive VAP & 27.84\% \\
  Positive IRI and negative VAP & 8.55\% \\
  Negative IRI and positive VAP & 6.78\% \\
  Negative IRI and negative VAP & 56.83\% \\
  \bottomrule
  \end{tabular}}
\end{table}

\end{document}